%% file: main.tex
\documentclass[10pt,twocolumn,letterpaper]{article}

\usepackage[pagenumbers]{cvpr} 
\usepackage[T1]{fontenc}

\definecolor{cvprblue}{rgb}{0.21,0.49,0.74}
\usepackage[pagebackref,breaklinks,colorlinks,allcolors=cvprblue]{hyperref}
\usepackage{multirow}
\def\paperID{887} 
\def\confName{CVPR}
\def\confYear{2025}

\title{Exploring Contextual Attribute Density in Referring Expression Counting}
\author{
Zhicheng Wang\quad Zhiyu Pan\thanks{Corresponding author.}\quad Zhan Peng\quad Jian Cheng\quad Liwen Xiao\quad Wei Jiang\quad Zhiguo Cao \\
School of AIA, Huazhong University of Science and Technology\\
{\tt\small \{zhicheng\_wang, zhiyupan, zgcao\}@hust.edu.cn}
}

\begin{document}
\maketitle
\input{sec/0_abstract}    
\input{sec/1_intro}

\input{sec/2_related}

\input{sec/3_method}
\input{sec/4_experiment}
\input{sec/5_conclusion}

{
    \small
    \bibliographystyle{ieeenat_fullname}
    \bibliography{main}
}

\input{sec/X_suppl}
\end{document}

%% file: sec/0_abstract.tex
\begin{abstract}

Referring expression counting (REC) algorithms are for more flexible and interactive counting ability across varied fine-grained text expressions. However, the requirement for fine-grained attribute understanding poses challenges for prior arts, as they struggle to accurately align attribute information with correct visual patterns. Given the proven importance of ``visual density'', it is presumed that the limitations of current REC approaches stem from an under-exploration of ``contextual attribute density'' (CAD). In the scope of REC, we define CAD as the measure of the information intensity of one certain fine-grained attribute in visual regions. To model the CAD, we propose a U-shape CAD estimator in which referring expression and multi-scale visual features from GroundingDINO can interact with each other. With additional density supervision, we can effectively encode CAD, which is subsequently decoded via a novel attention procedure with CAD-refined queries. Integrating all these contributions, our framework significantly outperforms state-of-the-art REC methods, achieves $30\%$ error reduction in counting metrics and a $10\%$ improvement in localization accuracy. The surprising results shed light on the significance of contextual attribute density for REC. Code will be at \href{https://github.com/Xu3XiWang/CAD-GD}{github.com/Xu3XiWang/CAD-GD}.

\end{abstract}

%% file: sec/1_intro.tex
\section{Introduction}
\label{sec:intro}

Object counting algorithms estimate the number of specific objects in query images. Early research has explored various forms of object descriptions, including class-specific~\cite{zhang2015cross, onoro2016towards}, class-agnostic~\cite{ranjan2021learning, shi2022represent, djukic2023low, pelhan2024dave}, exemplar-guided~\cite{ranjan2021learning}, and text-guided~\cite{xu2023zero, jiang2023clip} approaches. However, these methods lack the flexibility to distinguish between fine-grained, open-world descriptions, \eg, differentiating a \textit{walking person} from a \textit{person riding motorcycle}. To address this, recent work~\cite{dai2024referring} proposes the referring expression counting (REC) task which aims to count fine-grained and contextually nuanced objects within the same class.




\begin{figure}[tp]
    \centering
    \includegraphics[width=\linewidth]{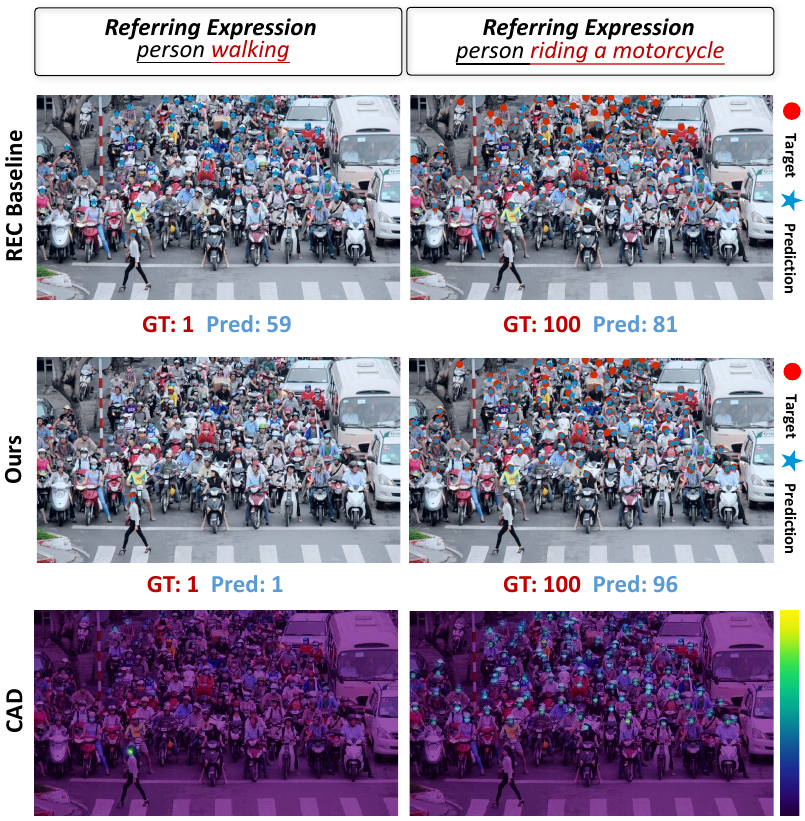}
    \caption{\textbf{The efficacy of contextual attribute density (CAD).} According to the qualitative comparison above, the REC baseline~\cite{dai2024referring} often over-counts when it focuses primarily on class information rather than fine-grained attributes, and it under-counts in cases of occlusion or scale variations. By modeling CAD, our method more accurately aligns fine-grained attributes with corresponding visual regions, effectively reducing counting errors.}
    \label{fig:comparasion}
\end{figure}

GroundingREC~\cite{dai2024referring} is firstly developed as the REC baseline with the power of the open-world detection model GroundingDINO~\cite{liu2023grounding}. However, distinguishing between objects of the same class based on nuanced attribute differences is challenging, which makes the REC baseline struggle to align attributes with corresponding visual regions.
As shown in Fig.~\ref{fig:comparasion}, two primary types of failure cases can be observed in the REC baseline: (1) Over-counting: the baseline may place excessive emphasis on class information while neglecting fine-grained attributes, \eg, \textit{walking}, leading to the inclusion of objects with wrong attributes; (2) Under-counting: the baseline might miss objects with the given attribute when occlusion and scale variations arise. 
These two failure cases may be due to the baseline's reliance on one-to-one matching between texts and image regions, a typical detection-to-count pipeline.
Many previous counting approaches~\cite{cao2018scale,idrees2018composition,cheng2022rethinking,hu2020count} have proved that visual density can provide information about scale-robust spatial distribution. However, in modern open-world perception models~\cite{liu2023grounding, ren2024grounding, lai2024lisa}, the capacity to perceive density for a specific attribute is often overlooked.
Inspired by the success of visual density modeling, we propose investigating whether modeling contextual attribute density (CAD) for an open-world perception model could help address the above issues in a cross-modal context.

In this work, we attempt to make the REC model aware of the CAD. Analogous to the concept of visual density, we define CAD as the information intensity measurement of one certain fine-grained attribute about one object conveyed in query image regions of different scales. Based on our definition, modeling CAD requires both the image and text information, with the image information encoded in a multi-scale format. Hence, we propose a CAD generate module with U-shape CAD estimator (CADE) that encodes the multi-layer CAD features. The CADE is provided with the visual features and similarity features between visual and text to estimates the CAD map for a referring expression, where the intermediate features serve as representations of CAD. After encoding the CAD, the next problem is how to facilitate the counting procedure with the CAD features. We fuse the CAD features with original visual features from GroundingDINO via both spatial and channel attention. In this way, spatial location and channel importance prior are provided for visual features for referring expression separately. Additionally, the counting queries in the localization decoder for counting are also dynamically initialized by both the text features and their corresponding CAD information. With this dynamic initialization module, the feature differentiation of the initial queries under different inputs is enhanced for better counting accuracy.

Integrating CAD into our REC framework leads to substantial improvements.
Experimental results on the public benchmark REC-$8$K~\cite{dai2024referring} show that our framework outperforms previous state-of-the-art REC baselines by large margins, achieving $37.8\%$ and $33.2\%$ relative error reductions on the validation and test sets, respectively. Besides, We also conduct experiments on the previous class-agnostic counting benchmark FSC-$147$~\cite{ranjan2021learning} with a zero-shot setting, with results further underscoring the effectiveness of our framework. These findings highlight the significance of CAD. Our contributions are three-fold:
\begin{itemize}
\item[$\bullet$] We pioneer the concept of contextual attribute density in REC, enabling more precise differentiation between fine-grained descriptions of objects within the same class.
\item[$\bullet$] We propose the contextual attribute density estimator and an attention procedure with a query enhancement, including the contextual attribute density into the REC pipeline.
\item[$\bullet$] We achieve significant performance improvements on public benchmarks, demonstrating the value of contextual attribute density modeling for REC.
\end{itemize}

%% file: sec/2_related.tex
\section{Related Work}
\label{sec:related}

The progression toward more generalized, fine-grained counting tasks has shifted from class-specific to class-agnostic methods, and also from visual to textual guidance. This section reviews prior works contributing to the development of open-world counting models, emphasizing advancements in visual density for counting tasks given our introduction of contextual attribute density.

\noindent\textbf{Advancing to open-world counting.} 
Counting tasks originally focus on specific categories, such as crowd~\cite{abousamra2021localization, cao2018scale, idrees2018composition}, vehicle~\cite{hsieh2017drone}, wheat head~\cite{madec2019ear, lu2017tasselnet}, and so on. These early methods are limited to predefined target types and struggle to generalize across diverse object categories. The class-agnostic counting (CAC)~\cite{lu2019class} marks a significant step toward more flexible counting approaches~\cite{shi2022represent,djukic2023low, wang2024vision, pelhan2024dave}. Defined as a template-matching problem~\cite{lu2019class}, CAC achieves generalized counting by using visual exemplars to count objects beyond specific classes. Notably, visual density-based methods~\cite{shi2022represent, djukic2023low, wang2024vision} consistently outperform traditional detection-based counting methods such as Mask-RCNN~\cite{he2017mask} and RetinaNet~\cite{ross2017focal} on counting benchmarks. However, the reliance on visual templates is still a notable limitation in practical applications. To address this issue, textual-based counting~\cite{xu2023zero,jiang2023clip} is proposed by leveraging the multi-modal ability of CLIP~\cite{radford2021learning}. Although incorporating visual guidance information has been proved beneficial~\cite{amini2024countgd}, researchers pursue a more generalized setting: with only text expressions, fine-grained, general-purpose objects are expected to be counted, which is referring expression counting (REC)~\cite{dai2024referring}. The REC baseline solely relies on the multi-modal localization ability of GroundingDINO~\cite{liu2023grounding}, and interpreting some fine-grained expressions accurately remains challenging, as shown in Fig.~\ref{fig:comparasion}.

\noindent\textbf{Density modeling in computer vision.} Beyond the task of counting, there are some works in other fields using visual density features to provide the spatial and target number prior, \eg, instance segmentation~\cite{cholakkal2019object} and tiny object detection~\cite{li2020density, huang2024dq, zhou2017accurate, duan2021coarse}. Cholakka \textit{et al.}~\cite{cholakkal2019object} use density maps to provide prior information on object counts and spatial locations, thereby assisting instance detection tasks under image-level supervision. Huang \textit{et al.}~\cite{huang2024dq} propose DQ-DETR for tiny object detection, which utilizes the density feature to provide the general target number and spatial prior for DETR~\cite{carion2020end} structure detection pipeline. In contrast to these methods, our approach focuses more on the role of contextual attribute density for fine-grained expression understanding in a multi-modal setting.

%% file: sec/3_method.tex
\begin{figure*}[tp]
    \centering
    \includegraphics[width=\linewidth]{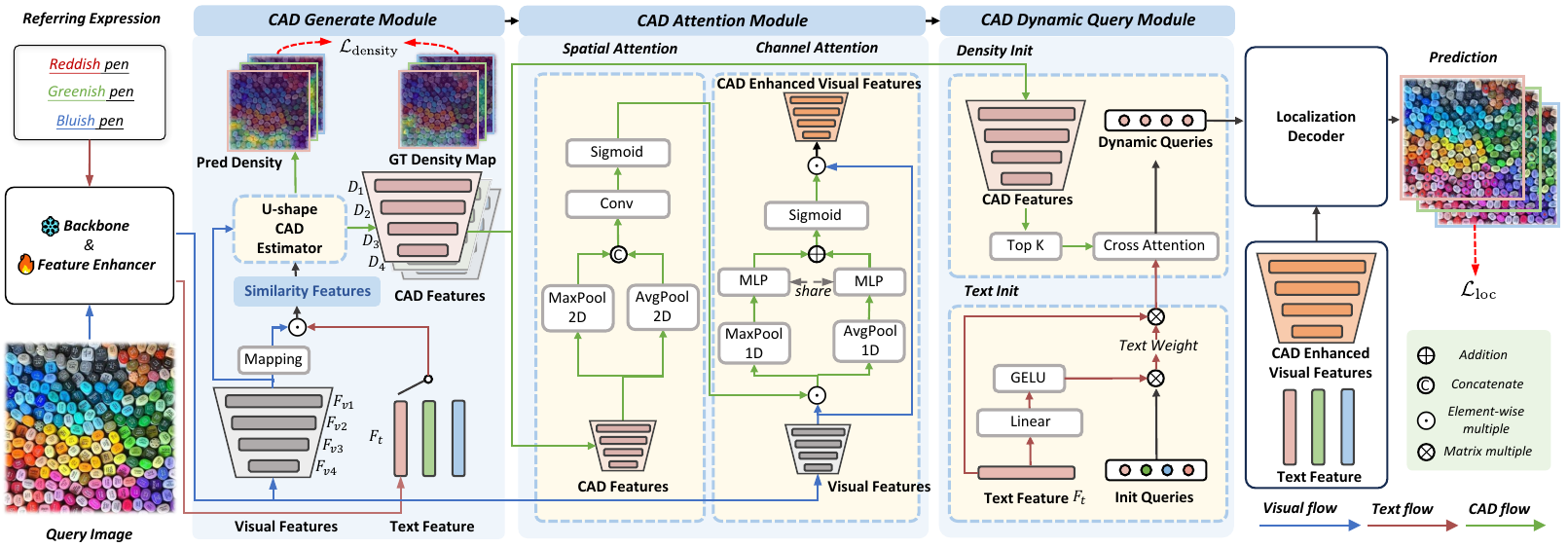}
    \vspace{-18pt}
    \caption{\textbf{The framework of Contextual Attribute Density Aware GroundingDINO (CAD-GD).} A query image and referring expression are sent into backbones separately together with the feature enhancer to get the visual features and the text feature. Then we obtain the Contextual Attribute Density (CAD) features by the CAD Generate Module supervised by the GT contextual attribute density map. To leverage the CAD information, we design the CAD Attention Module and CAD Dynamic Query Module to enhance visual features and query contents separately. Then we sent the dynamic queries, CAD enhanced visual features, and the text feature into the localization decoder to get the final localization prediction.}
    \label{fig:pipeline}
\end{figure*}


\section{Method}
\label{sec:method}

In this section, we first introduce the overview of our framework for Referring Expression Counting (REC), named Contextual Attribute Density Aware GroundingDINO (CAD-GD). The overview diagram of our approach is shown in Fig. \ref{fig:pipeline}. Given an input query image $I\in \mathbb{R}^{H\times W \times 3}$ and a referring expression text $T=\{w_i\}^{N}_{i=1}$ with $N$ words, our model predicts a localization counting result. 
For each $(I, T)$ pair, we first send 
multi-scale image features and text features extracted from the image and text backbones into feature enhancer~\cite{liu2023grounding} to obtain 
cross-modality visual features $\{F_{vi}\in \mathbb{R}^{h_i \times w_i \times C}\}_{i=1}^{4}$
and text feature $F_{t}\in \mathbb{R}^{N\times C}$, where the $C$ indicates the number of channels.

Afterwards, we feed visual features $\{F_{vi}\}_{i=1}^{4}$ and text feature $F_{t}$ into the U-shape contextual attribute density estimator (CADE) to obtain the contextual attribute density (CAD) features $\{D_i \in \mathbb{R}^{h_i \times w_i \times C}\}_{i=1}^{4}$. 
We then inject the CAD features $\{D_i\}_{i=1}^{4}$ into visual features $\{F_{vi}\}_{i=1}^{4}$ to enhance the visual feature with spatial and channel attention in CAD Attention Module. Besides, for the 
initialization of queries in the cross-modality localization decoder~\cite{liu2023grounding}, we selected the $K$ visual feature positions with the highest scores among them as the initial query locations by calculating the similarity between the visual and text features, while for the initial content features of the query, 
we dynamically initialize them with text features 
$F_{t}$ and CAD features $\{D_i\}_{i=1}^{4}$ corresponding to the query positions.
Then we refine the queries by the cross-modality decoder
at every layer 
with CAD enhanced visual 
and 
text features. 
The output queries of the last decoder layer will be used to predict the localization of referring expression objects.

\subsection{CAD Generation Module}


The first step to model CAD for referring expression counting is to obtain the CAD features, which should be responsive to input text with fine-grained attributes. 
To obtain the satisfying CAD features, 
we identify several requirements for obtaining satisfying CAD features: 
(a) the image-text similarity features are required to guide the estimator to focus on areas corresponding to the input referring expression; 
(b) due to the density diversity of targets in the counting task, a multi-scale structure is of benefit for the robustness of counting~\cite{idrees2013multi}.


Motivated by the above requirements, the CAD Generate Module is proposed, as in Fig.~\ref{fig:pipeline}. 
Specifically,
we first utilize the well-pretrained linear projector from GroundingDINO to project the visual features $\{F_{vi}\}_{i=1}^{4}$ into the text feature space. Then we calculate the similarity feature $\{S_{i}\in \mathbb{R}^{h_i \times w_i \times C}\}_{i=1}^{4}$ between the well-aligned visual feature $F_{vi}$ and corresponding referring expression feature $F_t$ as follow:
\begin{equation}
S_i =\mathrm{Proj}({F_v}_i) \cdot F_t,
\end{equation}
where the $\mathrm{Proj}$ consists of a single linear layer followed by a layer norm, and the $\cdot$ represents the element-wise multiplication. Then we send the $\{S_{i}\}_{i=1}^{4}$ together with the visual feature $\{F_{vi}\}_{i=1}^{4}$ into a U-shape CAD estimator (CADE) to estimate the CAD map for a referring expression, where the intermediate features serve as the CAD features $\{D_i\}_{i=1}^{4}$.

\noindent\textbf{Supervision Signal.}  The final output of CADE is the density map for referring expression. In order to guide the CADE to learn the contextual attribute density map, we adopt a conventional $\ell _2$ loss as the counting loss $\mathcal{L}_{\text{density}}$:  
\begin{equation}
\mathcal{L}_{\text{density}} = \left\| D_{\text{pred}}(I, T) - D_{\text{gt}}(I, T) \right\|_2^2,
\end{equation}
where $D_{\text{pred}}(I, T)$ and $D_{\text{gt}}(I, T)$ denote the prediction and ground truth contextual attribute density map 
respectively.

\subsection{CAD Attention Module}
After obtaining the CAD features $\{D_i\}_{i=1}^{4}$, the next core step is to consider how to inject the CAD information into the localization process, which 
follows 
a typically DETR decoder structure~\cite{liu2023grounding}. In the DETR-like structure, the visual features $\{F_{vi}\}_{i=1}^{4}$ are crucial for the localization process due to its cross-attention with queries. Motivated by the high relevance between the density feature and the attention map~\cite{wang2024vision}, 
we inject the CAD features into visual features by attention mechanism, as shown in Fig.~\ref{fig:pipeline}.




\noindent\textbf{Spatial Attention.} Since the visual features need to be matched with queries to find the target location, foreground enhancement for visual features on areas corresponding to the input referring expression can 
help to get accurate localization results. Hence we design the spatial attention mechanism to enhance the CAD corresponding foreground areas, utilizing distribution prior from CAD.
In specific, for the $i_{th}$ scale of $\{D_i\}_{i=1}^{4}$, we first apply the average pooling and max pooling for $D_{i}$ on the channel dimension. Then we concatenate these two channel pooled features together to get the spatial attention features $F_{i}^s\in \mathbb{R}^{h_i \times w_i \times 2}$. Afterward, the spatial attention features $F_{i}^s$ are sent into CNN layers followed by a sigmoid function to obtain the spatial attention map for $F_{vi}$. The process is as follows:
\begin{equation}
{A}_i^s=Sigmoid(Conv([\mathrm{Max}(D_i), \mathrm{Avg}(D_i)])),
\end{equation}
where ${A}_i^s\in \mathbb{R}^{h_i \times w_i \times 1}$ is the spatial attention map for ${F_v}_i$.
Then we multiple the visual feature ${F_v}_i$ and corresponding spatial attention map as follow:
\begin{equation}
{\dot {F_v}}_i = {F_v}_i \cdot  {A}_i^s.
\end{equation}
This results $\{\dot{F_{vi}}\}_{i=1}^4$ are the spatially attended visual features for each scale $i$.




\noindent\textbf{Channel Attention.} The previous work~\cite{shi2022represent} shows that channel attention can help to distinguish different class targets. For the referring expression counting task, it is even harder to distinguish targets due to the contextual attribute understanding requirement. So it is beneficial to make use of channel attention to enhance the distinguishing ability of the REC model. However, the previous channel attention in the CAC task neglects the scale variance due to its relation to fixed-size exemplars. Thanks to the multi-layer CAD features, our channel attention can vary on different scales. Specifically, for the $i_{th}$ scale, we first apply average pooling and max pooling followed by shared MLP on the spatial attention enhanced visual feature $\dot F_{vi}$ along the spatial dimension to get two channel attention features $F_i^c \in \mathbb{R}^{1 \times C}$. We add two features up, then feed the added feature into the sigmoid function to get the channel attention map ${A}_i^c$: 
\begin{equation}
{A}_i^c=Sigmoid(MLP(\mathrm Avg(\dot F_{vi})+\mathrm Max(\dot F_{vi}))).
\end{equation}

Afterwards, we multiple the spatial attention enhanced visual feature $\dot F_{vi}$ with channel attention map ${A}_i^c$ in every scale $i$ to obtained the final CAD enhanced visual features $\{\hat F_{vi}\}_{i=1}^{4}$ as follow:
\begin{equation}
{\hat {F_v}}_i =\dot {F_v}_i \cdot  {A}_i^c.
\end{equation}

\subsection{CAD Dynamic Query Module}


\begin{figure*}[tp]
    \centering
    \includegraphics[width=\linewidth]{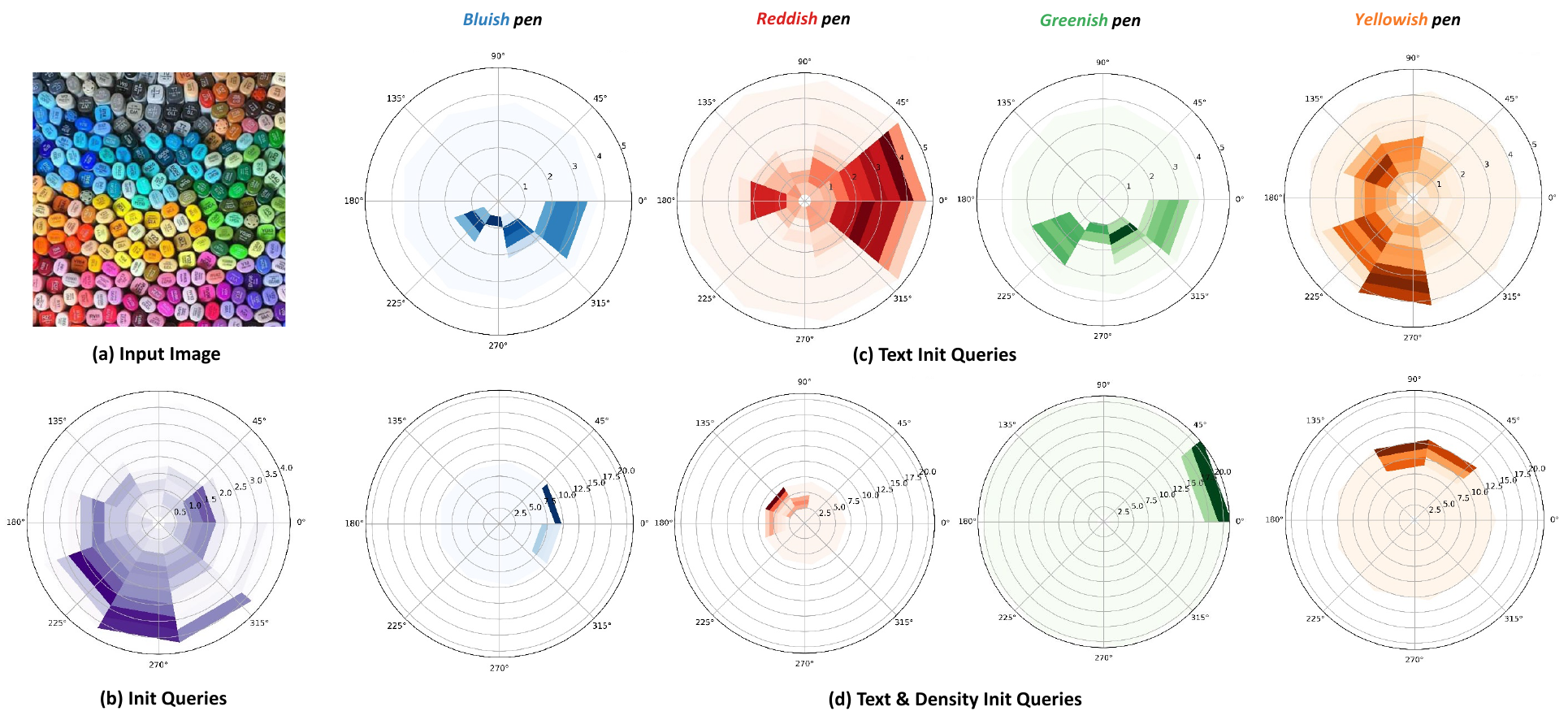}    \caption{\textbf{Visualization of queries.} We visualize the init queries $Q$, text dynamic initialized $\dot Q$ and the CAD dynamic initialized queries $\hat Q$ in the polar coordinate system by t-SNE~\cite{van2008visualizing}. (b) represent the feature distribution of init queries in the polar coordinate system, which are the same for different referring expressions. (c) represent the feature distribution of text init queries. As shown in (c), the distribution of ''bluish pen" and ''greenish pen" is quite similar even with much overlap. (d) represent the feature distribution of text \& density init queries. As shown in (d), the distribution of queries is quite different, which means they can be distinguished easily.}
    \label{fig:query}
\end{figure*}

The queries consisting of their locations and corresponding content features are another essential component for the localization decoder.
For the initial location of queries, we follow the GroundingDINO strategy of selecting visual feature positions with the top $K$ similarity scores to the text feature. For the corresponding content features $Q\in \mathbb{R}^{K \times  C}$ of queries which are learnt embedding in GroundingDINO, we argue that the fixed embedding is not the ideal initialization for query contents in REC. In the REC task, the target with a specific attribute is given, hence the initialized $K$ content is expected to contain as many kinds belonging to this given referring expression as possible.


\noindent\textbf{Text Init.} Therefore, as mentioned above, for the referring expression counting task, since it has only one given target text, we propose to dynamically initialize the query contents with text features first as shown in Fig.~\ref{fig:pipeline}.
In specific, we first multiply the input text feature $F_t$ with the learning matrix $M \in \mathbb{R}^{C \times C}$, then send it into the GELU function to obtain the dynamic text feature $T\in \mathbb{R}^{N \times C}$, where $N$ is the words number. Afterward, we multiply the dynamic text feature $T$ with init queries $Q\in \mathbb{R}^{K \times C}$, where $K$ is the query number. We can get the dynamic text weight map $W\in \mathbb{R}^{K \times N}$. Finally, we multiply the dynamic weight with the text feature $F$ to get the text-dynamic init queries $Q_t$. The process is as follows:
\begin{equation}
\dot Q = (Q\times(F_t\times M)^\top)\times F_t.
\end{equation}
Compared to the original queries, the meaning of the text-dynamic init queries obtained in this method is more explicit and can be understood as a combination of different words in the input referring expression.

\noindent\textbf{Density Init.} While using text combination-based initialization makes it possible for queries to contain different attribute weights, it completely ignores the variance of the given query image. 
Therefore, we propose to use CAD features which are highly related to the visual features of the given referring expression to further initialize query contents to improve the distinguishing ability.
Specifically, we first find $K$ CAD features $D_K$ corresponding to the query positions, which has the top $K$ similarity scores to the text feature. 
Then, we input the corresponding CAD feature $D_K$ as key and value and the text-dynamic queries $\dot Q$ as the query for a cross-attention layer to fuse. Finally, we can obtain the CAD dynamic initialized queries $\hat Q$.

In order to verify that the CAD Dynamic Query Module is conducive to distinguishing between different referring expressions. We visualize the init queries $Q$, text dynamic initialized $\dot Q$ and the CAD dynamic initialized queries $\hat Q$ in the polar coordinate system by t-SNE~\cite{van2008visualizing}. As shown in Fig.~\ref{fig:query}, for queries with only text initialization, it is hard to distinguish the bluish pen and greenish pen which are relatively close and similar in the feature space. However, with the help of CAD initialization, the query contents for different input referring expressions are distinguishable.

\subsection{Overall Objective}

Our loss function consists of two components, the first one is the localization loss $\mathcal{L}_{\text{loc}}$ as previous REC work based on GroundingDINO~\cite{dai2024referring}, and the second of which is the density loss $\mathcal{L}_{\text{density}}$. The more detail for $\mathcal{L}_{\text{loc}}$ is on the supplementary. Hence the final loss is a weighted sum of two losses:
\begin{equation}
\mathcal{L} = \mathcal{L}_{\text{loc}} + \alpha \cdot \mathcal{L}_{\text{density}}\,,
\end{equation}
where $\alpha$ is a parameter to control the influence of $\mathcal{L}_{\text{density}}$.


%% file: sec/4_experiment.tex
\section{Experiments}
\vspace{4pt}
\subsection{REC Datasets \& Metric}

\noindent\textbf{REC-8k~\cite{dai2024referring}.} REC-8K is the first REC task dataset, which contains 8011 different images and 17122 Image-RE pairs. The dataset is split into train, val, test sets which contain 10555, 3336, 3231 Image-RE pairs, 4923, 1566, 1522 images, and 723, 341, 299 REs respectively. 80 REs are shared between train and val sets and 79 REs are shared between train and test sets. The original REC-8K dataset does not contain ground truth contextual attribute density maps, so we generate these maps using a fixed-size Gaussian kernel of 15, following the previous method~\cite{lempitsky2010learning}.



\noindent\textbf{Metrics.} Following prior work on object counting, the Mean Absolute Error (MAE) and the Root Mean Squared Error (RMSE) are used to measure counting performance. For the REC-8K dataset, we include Precision, Recall, and F1 scores as metrics for localization as GroundingREC~\cite{dai2024referring}.

\definecolor{lightred}{RGB}{239,245,251}  
\begin{table*}[t]
  \centering
  \renewcommand{\arraystretch}{1}
  \addtolength{\tabcolsep}{-1.5pt}
\caption{\textbf{Comparison with the state-of-the-art approaches on the REC-8K dataset.} The best performance is \textbf{boldfaced}, and the second is \underline{underlined}. GroundingREC* is reproduced following the training setting in paper~\cite{dai2024referring} using swin-b as the visual backbone because they do not report the result. Results with \dag{} indicate that we use the number of density map estimates to determine the number of queries selected for localization.}
\begin{tabular}{@{}lccccccccccc@{}}
\toprule
\multirow{2}{*}{Method}            & \multirow{2}{*}{Backbone}              & \multicolumn{5}{c}{Val set}                         & \multicolumn{5}{c}{Test set}       \\ \cmidrule(lr){3-7}  \cmidrule(lr){8-12}
                  &                       & MAE~$\downarrow$            & RMSE~$\downarrow$           & Prec~$\uparrow$ & Rec~$\uparrow$  & F1~$\uparrow$   & MAE~$\downarrow$   & RMSE~$\downarrow$  & Prec~$\uparrow$ & Rec~$\uparrow$  & F1$\uparrow$   \\ \hline
Mean              &- & 14.28         & 27.75          & -    & -    & -    & 13.75 & 25.91 & -    & -    & -    \\
ZSC~\cite{xu2023zero}               & Res-50             & 14.84         & 31.30          & -    & -    & -    & 14.93 & 29.72 & -    & -    & -    \\
ZSC~\cite{xu2023zero}               & Swin-T                & 12.96         & 26.74          & -    & -    & -    & 13.00 & 29.07 & -    & -    & -    \\
CounTX~\cite{amini2023open}            & ViT-B              & 11.88         & 27.04          & -    & -    & -    & 11.84 & 25.62 & -    & -    & -    \\
GroundingDINO~\cite{liu2023grounding}     & Swin-T                & 9.03          & 21.98          & 0.56 & \underline{0.76} & 0.65 & 8.88  & 21.95 & 0.59 & \underline{0.76} & 0.66 \\
GroundingREC~\cite{dai2024referring}      & Swin-T                & 6.80          & 18.13          & 0.65 & 0.71 & 0.68 & 6.50  & 19.79 & 0.67 & 0.72 & 0.69 \\
\rowcolor{lightred} CAD-GD (ours) & Swin-T                & 5.43          & 15.01          & 0.68 & 0.72 & 0.70 & 5.29  & 17.08 & 0.71 & 0.73 & 0.72 \\
\rowcolor{lightred} CAD-GD\dag{} (ours) & Swin-T                &4.58         & \underline{13.24}          & 0.68 & 0.71 & 0.70 & \underline{4.59}  & 14.68 & 0.72 & 0.70 & 0.71 \\ \midrule
GroundingREC*~\cite{dai2024referring}      & Swin-B                & 5.66          & 15.24          & 0.66 & \textbf{0.77} & 0.71 & 5.42  & 18.47 & 0.71 & 0.69 & 0.70 \\
\rowcolor{lightred}CAD-GD (ours) & Swin-B                & 4.83          & 13.52          & \underline{0.74} & \underline{0.76} & \textbf{0.75} & 4.94  & \underline{14.65} & \underline{0.75} & \textbf{0.77} & \textbf{0.76} \\
\rowcolor{lightred}CAD-GD\dag{} (ours) & Swin-B                & \textbf{4.23} & \textbf{13.14} & \textbf{0.76} & 0.70 & \underline{0.73} & \textbf{4.34}  & \textbf{12.93} & \textbf{0.77} & 0.71 & \underline{0.74} \\ \bottomrule
\end{tabular}

\label{tab:rec8k}
\end{table*} 

\vspace{4pt}

\subsection{Implementation Details}

Our model is built on the GroundingDINO~\cite{liu2023grounding} where the text and visual backbone, feature enhancer, and localization decoder are from. We provide two versions of our model, trained with Swin-T and Swin-B as visual backbone separately. Please refer to the supplementary for more details about network architectures.



\noindent\textbf{Training.} We freeze the visual and text backbone to keep the feature space aligned. For the REC-8K dataset, we use the AdamW as the optimizer to train the model for 20 epochs. The learning rate is set to $1e-5$ and decays by $10\times$ on the $10^{th}$ epoch. For each batch, we send one image with multiple corresponding referring expressions. Besides, data augmentations are not used.



\noindent\textbf{Inference.} For the inference stage, we provide two strategies. In the first strategy, we use a fixed threshold following GroundingREC~\cite{dai2024referring}. For each output query, we compare its score with the CLS token by a threshold of 0.25 and scores with the rest of the text tokens by a threshold of 0.35. If all scores are higher than the threshold, we include the query in the final count. Eventually, the total number of queries that satisfy the threshold condition is considered to be the number of targets in the query image. The second strategy is to discard the threshold and use the target number estimated from the density map as the predicted number of outcomes $N$. Then we find the $N$ queries among all queries that have the highest scores with the CLS token and add them to the predicted outcomes. We use \dag{} to represent the model with the second strategy on Table~\ref{tab:rec8k}.

\subsection{Comparison with State of the Art}

\textbf{Quantitative Results.} 
In this section, we compare our CAD-GD with language-based counting methods, open-set detection model GroundingDINO, and state-of-the-art REC model GroundingREC on REC-8K datasets. Most of the results for comparison are obtained from previous paper~\cite{dai2024referring} directly. For the GroundingREC, we train a Swin-B version of GroundingREC, which is marked with * on the Table~\ref{tab:rec8k}, following the training setting in GroundingREC~\cite{dai2024referring} because it only has the swin-T version model.

As shown in Table~\ref{tab:rec8k}, for the Swin-T backbone model, CAD-GD\dag{} achieve a relative improvement of 32.6\% \textit{w.r.t} validation MAE and 29.4\% \textit{w.r.t} test MAE compared with GroundingDINO. Besides, our method reduces the validation RMSE by 27.0\% and test RMSE by 25.8\%. The reduction in the RMSE reflects the improved robustness of our method for query images of different densities.
For the localization metrics, our model improves precision consistently for localization, while the recall between GroundingREC and our model is similar. This phenomenon may be due to that the goal of our approach is to enhance the capture of contextual attributes for referring expression, thus improving the accuracy of localization instead of recall.

We also give in the table the performance of the CAD-GD with Swin-B as the visual backbone. As can be seen from Table~\ref{tab:ablation}, our method maintains a steady increase in counting accuracy compared to GroundingREC with Swin-B visual backbone. Meanwhile, as the scale of the visual backbone increases, our method can better improve the localization accuracy compared to GroundingREC. The reason may lie in that the quality of contextual attribute density is highly related to the quality of visual features.

\begin{figure*}[tp]
    \centering
    \includegraphics[width=\linewidth]{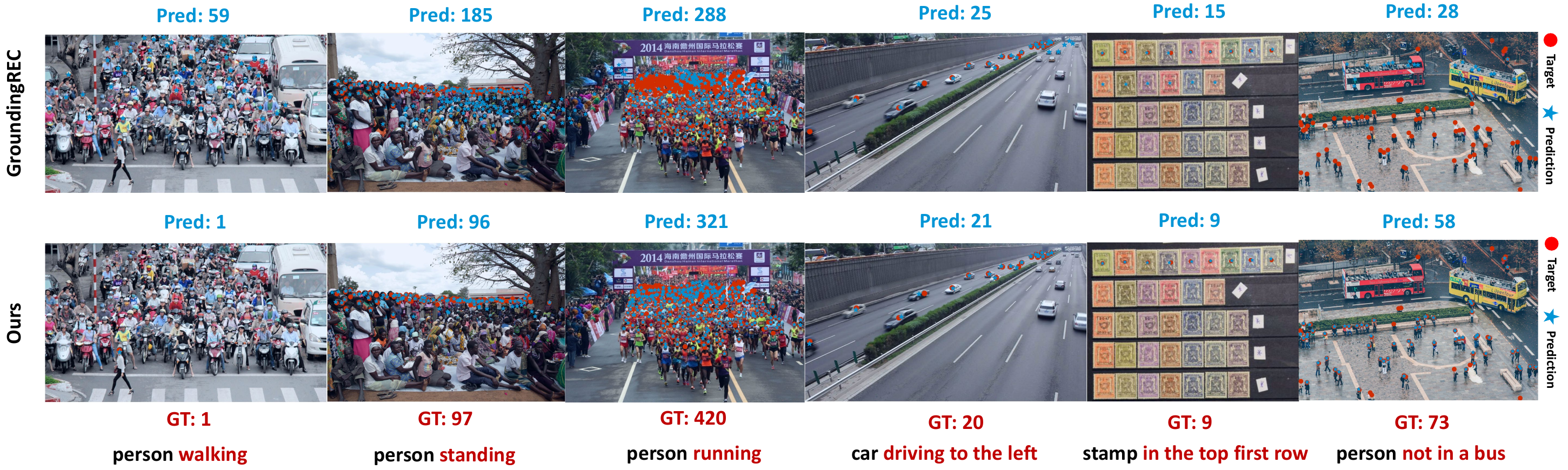}
    \vspace{-16pt}
    \caption{\textbf{Qualitative results on the REC-8k dataset.} Columns 1-3 contain examples with direct attributes for REs, and columns 4-6 contain those with context-related attributes. Our method consistently outperforms the GroundingREC with more precious locations.}
    \label{fig:qualitative}
\end{figure*}



\noindent\textbf{Qualitative Analysis.} For qualitative analysis, we compare our Density-GD with the state-of-the-art REC method GroundingREC in various scenes as shown in Fig.~\ref{fig:qualitative}. According to the difficulty of the input referring expression, we visualized examples with direct attributes (columns 1-3) and context-related attributes (columns 4-6), respectively. Compared to GroundingREC, even in different density environments, our method can better capture the attribute information of the right target, thus obtaining more accurate counting and localization results. Meanwhile, as shown in columns 4-6, our method can still achieve accurate counting results for referring expressions with contextual attributes. From the image corresponding to ``person not in a bus”, we can see that although the previous method can capture the two key information of ``person" and ``bus", it can not process the whole semantic features well, and thus wrongly count ``person in a bus". However, our approach can capture the context-dependent attribute correctly.

\begin{figure*}[tp]
    \centering
    \includegraphics[width=\linewidth]{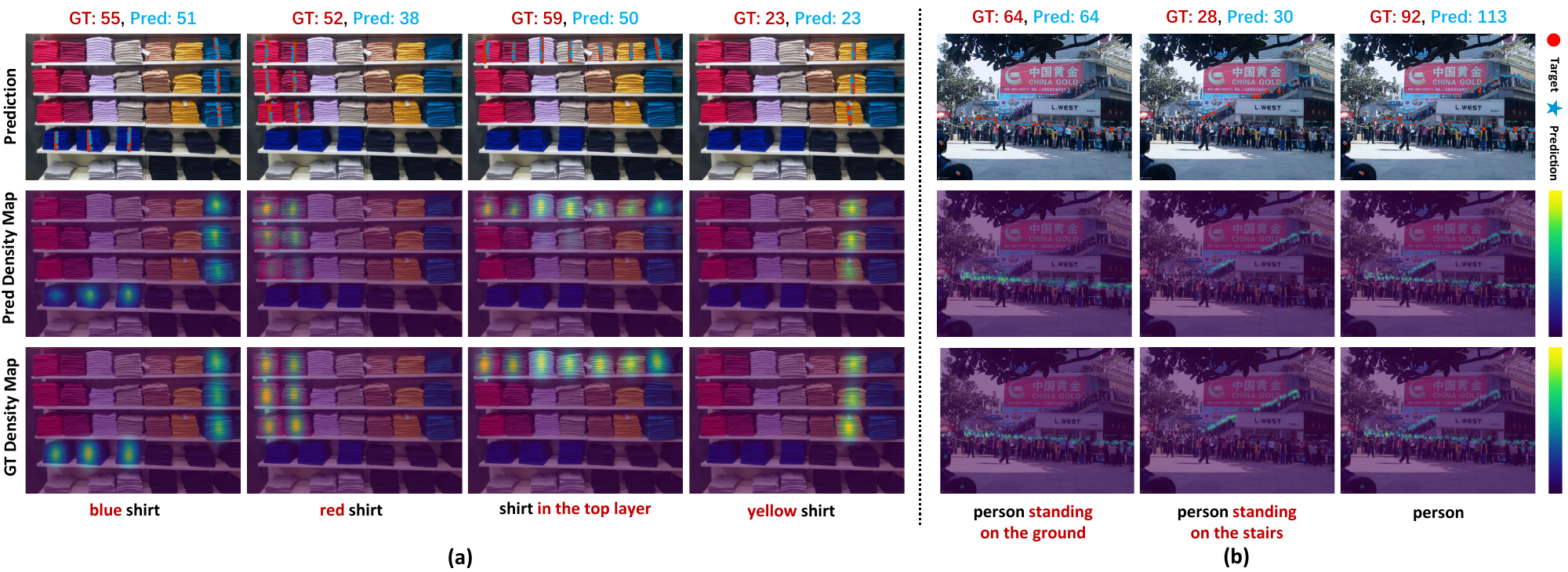}
    \vspace{-20pt}
    \caption{\textbf{Visualization of contextual attribute density map.} The predicted density map can capture the locations of referring expressions.}
    \label{fig:contexual_attribute_density}
\end{figure*}

\noindent\textbf{Effect of CAD.} To demonstrate the relation between the contextual attribute density and the final prediction results, we visualize the contextual attribute density map. As shown in Fig.~\ref{fig:contexual_attribute_density}, for the same input query image with different referring expressions, our contextual attribute density map can capture where the corresponding semantic features are located. As shown in the images in (a), the contextual attribute density map can distinguish different attributes on both appearance and spatial aspects. One can observe in (b) that our method can perform well even without any attributes for the target class.

\subsection{Ablation Study}

\begin{table}[]
\centering
  \addtolength{\tabcolsep}{-4.3pt}
\caption{\textbf{Ablation study on REC-8k dataset.} A represents the CAD Generate Module. B represents the CAD Attention Module, where the B1 and B2 represent the spatial attention and channel attention separately. C represents the CAD Dynamic Query Module, where C1 and C2 represent the text and density feature init. D represents the second strategy for inference, which is to use the target number estimated from the density map as the predicted number.}
\begin{tabular}{@{}cccccccccccc@{}}
\toprule
\multirow{2}{*}{No.} & \multirow{2}{*}{A} & \multicolumn{2}{c}{B} & \multicolumn{2}{c}{C} & \multirow{2}{*}{D} & \multicolumn{5}{c}{Val}  \\ \cmidrule(lr){3-4} \cmidrule(lr){5-6} \cmidrule(lr){8-12} 
                     &                       & B1          & B2         & C1          & C2          &                     & MAE   & RMSE   & Prec   & Recall & F1    \\ \midrule
R1                   &                       &             &            &             &             &                     & 6.52  & 17.72  & 0.632  & 0.703  & 0.665 \\
R2                   & \checkmark                     &             &            &             &             &                     & 6.17  & 16.38  & 0.641  & 0.709  & 0.673 \\
R3                   & \checkmark                     & \checkmark           &            &             &             &                     & 5.88  & 16.43  & 0.662  & 0.722  & 0.691 \\
R4                   & \checkmark                     & \checkmark           & \checkmark          &             &             &                     & 5.61  & 16.28  & 0.670  & 0.710  & 0.690 \\
R5                   & \checkmark                     & \checkmark           & \checkmark          & \checkmark           &             &                     & 5.67  & 14.43  & 0.678  & 0.703  & 0.690 \\
R6                   & \checkmark                     & \checkmark           & \checkmark          & \checkmark           & \checkmark           &                     & 5.43  & 15.01  & 0.683  & 0.718  & 0.700 \\
R7                   & \checkmark                     & \checkmark           & \checkmark          & \checkmark           & \checkmark           & \checkmark                   & 4.83  & 13.52  & 0.679  & 0.711  & 0.695 \\ \bottomrule
\end{tabular}

\label{tab:ablation}
\end{table}

We conduct the ablation study on REC-8K and provide quantitative results in Table~\ref{tab:ablation}. The plain version of CAD-GD is employed as the baseline which only consists of the standard backbone, feature enhancer, and localization decoder as GroundingREC~\cite{dai2024referring}.

\noindent\textbf{CAD Generate Module Effect.} By comparing R1 and R2 in Table~\ref{tab:ablation}, one can observe that SDCM can introduce a relative improvement of 5.4\% \textit{w.r.t.} MAE on the validation set. This performance boost indicates that the counting and localization ability of the model can be improved even by implicitly introducing contextual attribute density features.

\noindent\textbf{CAD Attention Module Effect.} The comparison of R2 ~\textit{vs.} R4 demonstrates that the SDAM improves the validation MAE by 0.51 and the precision by 0.29. This result suggests that the counting accuracy of the model can be improved by introducing CAD to visual features directly through SDAM.

\noindent\textbf{CAD Dynamic Query Module Effect.} By comparing R4 and R6 in Table~\ref{tab:ablation}, one can observe that both the MAE and RMSE are improved. However, the comparison of R4 and R5 demonstrates that the validation MAE declines slightly by only adding text init. The reason may lie in that only text init will reduce the perception of the query image.

\noindent\textbf{Impact of Inference Strategy.} By comparing R6 and R7 in Table~\ref{tab:ablation}, it can be observed that the TA can bring a relative improvement of 11.05\% \textit{w.r.t} MAE and 9.93\% \textit{w.r.t} RMSE on validation set. However, the performance on localization metrics even declines slightly. The phenomenon may lie in mismatch localization queries and density maps.

\subsection{Results on Zero-shot Counting}
\begin{table}[]
\centering
  \renewcommand{\arraystretch}{0.9}
  \caption{\textbf{Comparison with the state-of-the-art approaches on the FSC-147 dataset.} The upper and lower part of the table presents the results using visual exemplars and class text as guidance separately.}
\begin{tabular}{@{}lcccc@{}}
\toprule
\multirow{2}{*}{Method}           & \multicolumn{2}{c}{Val} & \multicolumn{2}{c}{Test}           \\\cmidrule(lr){2-3} \cmidrule(lr){4-5} 
                    & MAE        & RMSE       & MAE   & RMSE                       \\ \midrule
CounTR~\cite{liu2022countr}          & 13.13      & 49.83      & 11.95 & 91.23                      \\
LOCA~\cite{djukic2023low}            & 10.24      & 32.56      & 10.79 & 56.97                      \\
CACViT~\cite{wang2024vision}            & 10.63      & 37.95      & 9.13 & 48.96                      \\
DAVE~\cite{pelhan2024dave}            & 8.91       & 28.08      & 8.66  & 32.36                      \\
CountGD~\cite{wang2024vision}             & 7.10      & 26.08      & 5.74 & 24.09                      \\ \midrule
Patch-selection~\cite{xu2023zero} & 26.93   & 88.63          & 22.09 & 115.17                     \\
CLIP-count~\cite{jiang2023clip}      & 18.79      & 61.18      & 17.78 & 106.62 \\
VLCounter~\cite{kang2024vlcounter}       & 18.06      & 65.13      & 17.05 & 106.16                     \\
CounTX~\cite{amini2023open}          & 17.10      & 65.61      & 15.88 & 106.29                     \\
DAVE~\cite{pelhan2024dave}            & 15.48      & 52.57      & 14.90 & 103.42                     \\
GroundingREC~\cite{dai2024referring}    & 10.06      & 58.62      & 10.12 & 107.19                     \\
CountGD~\cite{amini2024countgd}         & 12.14      & 47.51      & 12.98 & 98.35                      \\ \midrule
CAD-GD (ours)               &  9.30       & 40.96      & 10.35 & 86.88                      \\ \bottomrule
\end{tabular}

\label{tab:fsc147}
\end{table}

\begin{table}[]
  \centering
  \addtolength{\tabcolsep}{-2pt}
    \renewcommand{\arraystretch}{0.9}
\caption{\textbf{Comparison with the state-of-the-art Sapproaches on the CARPK dataset.} Tuned means the method is finetuned on the CARPK training dataset.}
\begin{tabular}{@{}lcccc@{}}
\toprule
\multirow{2}{*}{Method}     & \multirow{2}{*}{Prompt}        & \multirow{2}{*}{Tuned} & \multicolumn{2}{c}{Test} \\ \cmidrule(lr){4-5} 
           &                &            & MAE         & RMSE       \\ \midrule
CLIP-count~\cite{jiang2023clip} & Text           & ×          & 11.96       & 16.61      \\
CounTX~\cite{amini2023open}     & Text           & \checkmark          & 8.13        & 10.87      \\
VLCounter~\cite{kang2024vlcounter}  & Text           & ×          & 6.46        & 8.68       \\
CACViT~\cite{wang2024vision}     & Visual         & \checkmark          & 4.91        & 6.49       \\
CountGD~\cite{amini2024countgd}    & Text           & ×          & 3.83        & 5.41       \\
CountGD~\cite{amini2024countgd}    & Text \& Visual & ×          & 3.68        & 5.17       \\
CAD-GD (ours)       & Text           & ×          & 3.29        & 4.56       \\ \bottomrule

\end{tabular}

\label{tab:carpk}
\end{table}

\label{sec:experiment}

\noindent\textbf{FSC-147.} Besides the REC task, we also evaluate our method on class agnostic counting to show our method is generalizable to prior tasks. The quantitative results are presented in Table~\ref{tab:fsc147}. We evaluate the benchmark dataset FSC-147~\cite{ranjan2021learning} in a zero-shot manner, which means the model is trained on the FSC-147 train set and evaluated on the val and test set of novel classes. For training and inference, we use the same training and testing strategy as CountGD~\cite{amini2024countgd}. Please refer to the supplementary material for more training and testing details.

As shown in Table~\ref{tab:fsc147}, our method achieves significantly lower counting errors than all prior zero-shot approaches. By comparing with the CAC methods, it can be seen that although our method only uses text as prompts, the counting error is still comparable with the method using extra visual exemplars as guidance.


\noindent\textbf{CARPK.} To test cross-dataset generalization, we test our method on the CARPK~\cite{hsieh2017drone} car counting dataset which is only trained on FSC-147 dataset. As shown in Table~\ref{tab:carpk}, CAD-GD achieves lower counting errors than other methods even with visual exemplars as guidance.

\vspace{-3pt}

%% file: sec/5_conclusion.tex
\section{Conclusion}



In this work, we study why the previous referring expression counting (REC) methods cannot well distinguish between fine-grained attributes for one class. The core issue is that previous methods overlook the contextual attribute density (CAD) conveyed in visual regions. Therefore, we propose the CAD-GD, in which the CAD information is modeled by a CAD estimator and injected into visual features and decoder queries via a well-designed CAD attention module and a CAD dynamic query module. By leveraging the CAD information, consistent performance improvements can be observed on varied benchmarks, which demonstrates the significance of CAD. We hope this work can inspire researchers to explore the CAD in other multi-modal tasks.

\label{sec:conclusion}

%% file: sec/X_suppl.tex
\clearpage
\setcounter{page}{1}
\maketitlesupplementary

\setcounter{section}{0}  
\renewcommand{\thesection}{S\arabic{section}}  

\section{Introduction}

This supplementary material includes the following contents:

\begin{itemize}
\item Detailed network architectures and implementation of CAD-GD;
\item Training and inference detail on FSC-147 and CARPK datasets;
\item Parameters and training resource; 
\item Additional qualitative results on REC-8K;
\item Limitations of CAD-GD.
\end{itemize}

\section{Network Architectures}
Here we detail some modules within our network.

\subsection{Backbone and Feature Enhancer}

The backbones consist of a visual backbone and a text backbone. 
The visual backbone is a Swin Transformer~\cite{liu2021swin}, we extract four different scales image features, which are $\frac{1}{4},\frac{1}{8},\frac{1}{16},\frac{1}{32}$ of input width and height, with the visual backbone. Then we sent these image features into $1\times 1$ convolutions to project into $256$ dimensions. 
The text backbone is a BERT-based text transformer~\cite{devlin2018bert}. The text backbone maps the input referring expression $T$ to a sequence of at most $256$ tokens. The encoded text tokens are $256$ dimensional feature vectors. 
We feed the features from the visual backbone and the text backbone into the feature enhancer from GroundingDINO~\cite{liu2023grounding}, composed of $6$ blocks. Each block uses a deformable self-attention~\cite{zhu2020deformable} to enhance the image features and a vanilla self-attention for text feature enhancement. Besides the self-attention mechanism, it contains an image-to-text cross-attention module and a text-to-image cross-attention module for feature fusion. After the feature enhancer, we can obtain the visual features $\{F_{vi}\}_{i=1}^{4}$ and text feature $F_{t}$.

\subsection{U-shape CAD Estimator}

\begin{figure}[tp]
    \centering
    \includegraphics[width=\linewidth]{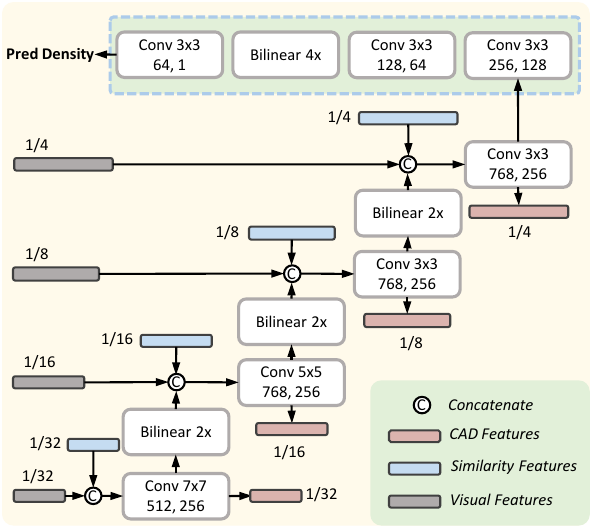}
    \caption{\textbf{U-shape CAD Estimator.} The visual features and the similarity features are sent into the U-shape CAD Estimator to obtain the CAD features. Each convolution block, in which the left and right dimensions are input and output dimensions separately, is followed by a ReLU as the activation function.}
    \label{fig:estimator}
\end{figure}

Here we detail the architecture of the U-shape CAD Estimator. As Figure~\ref{fig:estimator} shows, the U-shape CAD Estimator consists of layers of convolutions and bilinear upsampling. The intermediate features serve as CAD features.

\subsection{Localization Decoder}

The number of decoder queries is set to 900 as in GroundingDINO~\cite{liu2023grounding}. The localization decoder consists of 6 layers of decoder blocks. Each decoder block contains a self-attention layer for decoder queries, a visual cross-attention layer to combine CAD enhanced visual features, a text cross-attention layer to combine text features, and an FFN layer.

\subsection{Localization Loss}

The localization loss is the same as the loss in GroundingREC~\cite{dai2024referring}, which is used to optimize the location and classification of predicted points using ground truth points. By Bipartite Matching, we can first match the predicted points with the ground truth points, and then find the corresponding image tokens associated with the predicted points. The localization loss consists of a matching loss, a cross-entropy classification loss, and a contrastive loss.

\noindent\textbf{Matching loss.} The matching loss is for point regression which is the $L_1$ distance between the predicted points $\hat p_k$ and the ground truth points $p_k$. The total number of matched points is $K$. The $\mathcal{L}_{\text{match}}$ is calculated as follow:
\begin{equation}
\mathcal{L}_{\text{match}} = \frac{1}{K} \sum^{K}_{k=1}\left\| \hat p_k - p_k \right\|_1.
\end{equation}

\noindent\textbf{Cross-entropy classification loss.} Cross-entropy classification loss is for point classification, where $y_i$ is the ground truth label for $i^{th}$ text token and $\hat y_i$ is the class logit between $i^{th}$ text token and $k^{th}$ predicted image token. The loss is calculated by taking the mean of all scores for $N$ text tokens and then averaging over $K$ matched points.
\begin{small}
\begin{equation}
\mathcal{L}_{\text{cls}} = \frac{1}{K} \sum^{K}_{k=1}[-\frac{1}{N}\sum^{N}_{i=1}[y_{i}\mathrm{log}\hat y_i^k+(1-y_i)\mathrm{log}(1-\hat y_i^k)]].
\end{equation}
\end{small}

\noindent\textbf{Contrastive loss.} Contrastive loss is for image-text alignment. 
For the REC-8K, we take multiple inputs of different attributes of the same class. For the $k^{th}$ matched image token, we take the corresponding attributes of the input referring expression as the positive text sample, and the mean of other attributes of the same class as the negative sample. Then we can obtain the positive score $s^p_k$ and negative similarity score $s^n_k$ by calculating the similarity of the $k^{th}$ matched image token with the positive sample and the negative sample separately. The final contrastive loss for all the matched image tokens is as follows:
\begin{equation}
\mathcal{L}_{\text{constrast}} = - \frac{1}{K} \sum^{K}_{k=1}[\mathrm{log}(s^p_k)+\mathrm{log}(1-s^n_k)].
\end{equation}

Note that in the FSC-147~\cite{ranjan2021learning} dataset, we take classes different from the corresponding class of the $k^{th}$ matched image token in a batch as negative samples, because FSC-147 only contains the classes without exact attributes.

The final localization function is as follows:
\begin{equation}
\mathcal{L}_{\text{loc}} = \mathcal{L}_{\text{match}} +\lambda_1 \mathcal{L}_{\text{cls}} + \lambda_2 \mathcal{L}_{\text{constrast}},
\end{equation}
where the $\lambda_1$ and $\lambda_2$ are set to 5 and 0.06 respectively.

\subsection{Additional Implementation Detail}



The mapping function in CAD Generate Module is a linear layer with 256 input features and 256 output features followed with a LayerNorm.
The kernel size of the convolution layer in spatial attention is $7\times 7$. The MLP in channel attention consists of 4 layers of linear with 256 input and output features.
For the density init in CAD dynamic query module, we use a layer of cross-attention with a hidden dimension of 256 and 8 heads to fuse features. The density loss weight $\alpha$ is set to 1.

\section{Training and Inference}

Here we demonstrate the detailed training and inference setting for FSC-147 and CARPK datasets.

\subsection{FSC-147 Dataset}
\noindent\textbf{Training.} For the FSC-147~\cite{ranjan2021learning} dataset, CAD-GD is trained for 30 epochs using AdamW as the optimizer with a batch size of 4. The learning rate is set to $1e-5$ and decays by $10\times$ on the $15^{th}$ epoch. The visual backbone and text backbone are frozen during training. For the data augmentation, the minimum side length of the image is resized to a side length in \{480, 512, 544, 576, 608, 640, 672, 704, 736, 768, 800\} such that the aspect ratio of the image is maintained as in CountGD~\cite{amini2024countgd}. Following CountGD, all the classes in the FSC-147 training set are concatenated into a single caption with ``.". The density maps for training are directly from the FSC-147 dataset.

\noindent\textbf{Inference.} At inference, each image is resized such that its shortest side length is 800 pixels, and its aspect ratio is maintained. The image is then normalized and passed to the model. We use the adaptive cropping strategy in CountGD to overcome the 900 counting quota of the model. In specific, when the prediction number is larger than 600, we then crop the image into 4 pieces without overlapping. Then we resize all these pieces with their shortest side length to 800 pixels. To obtain the final count, the number of detected instances in each crop window are added together. The threshold of FSC-147 is set to 0.3.

\subsection{CARPK Dataset}
\noindent\textbf{Training.} We do not train CAD-GD on the CARPK~\cite{hsieh2017drone} dataset to validate the cross-dataset generalization ability. 

\noindent\textbf{Inference.} At inference, we do not use the resize and cropping strategy as in the FSC-147 dataset. For each image, we normalize it and then send it into the model for prediction. The threshold of CARKPK is set to 0.15.

\section{Parameters and Training Resource}
To verify the efficiency of our method, we compare our model size and floating point operations (FLOPs) with GroundingREC as shown in Table~\ref{tab:parameter}. The full training of CAD-GD with Swin-b visual backbone takes about 22GB on 1 Nvidia A100 GPU for about 20 hours.

\begin{table}[]
  \centering
  \addtolength{\tabcolsep}{-5pt}
    \renewcommand{\arraystretch}{1}
\caption{\textbf{Comparison of the model size and FLOPs.} We obtain the FLOPs using a $384\times 384$ image as input.}
\begin{tabular}{@{}lccc@{}}
\toprule
Method     & Backbone & Model Size (M)        & GFLOPs \\  \midrule
GroundingREC~\cite{dai2024referring} &Swin-T & 144.1           & 67.6              \\
CAD-GD    &Swin-T & 159.6           & 74.5                \\  \midrule
GroundingREC~\cite{dai2024referring} &Swin-B &  204.0          & 98.5                 \\
CAD-GD    &Swin-B &  219.5        & 105.4                \\ \bottomrule

\end{tabular}

\label{tab:parameter}
\end{table}

\section{Additional Qualitative Results}
\begin{figure*}[t!]
    \centering
    \includegraphics[width=0.99\linewidth]{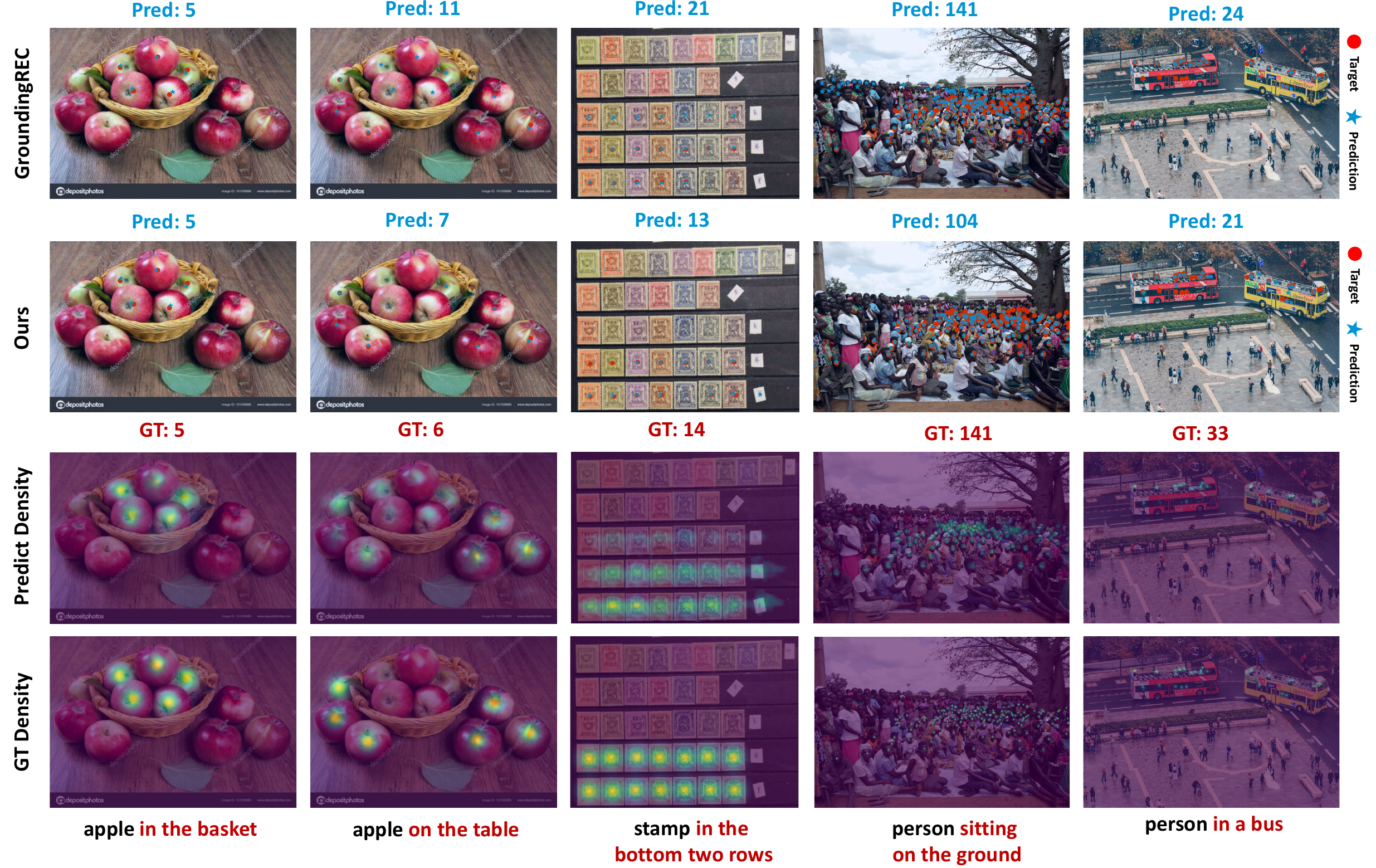}
    \vspace{-8pt}
    \caption{\textbf{Additional Qualitative Results 1.}}
    \label{fig:qualitative1}
\end{figure*}

\begin{figure*}[t!]
    \centering
    \includegraphics[width=0.99\linewidth]{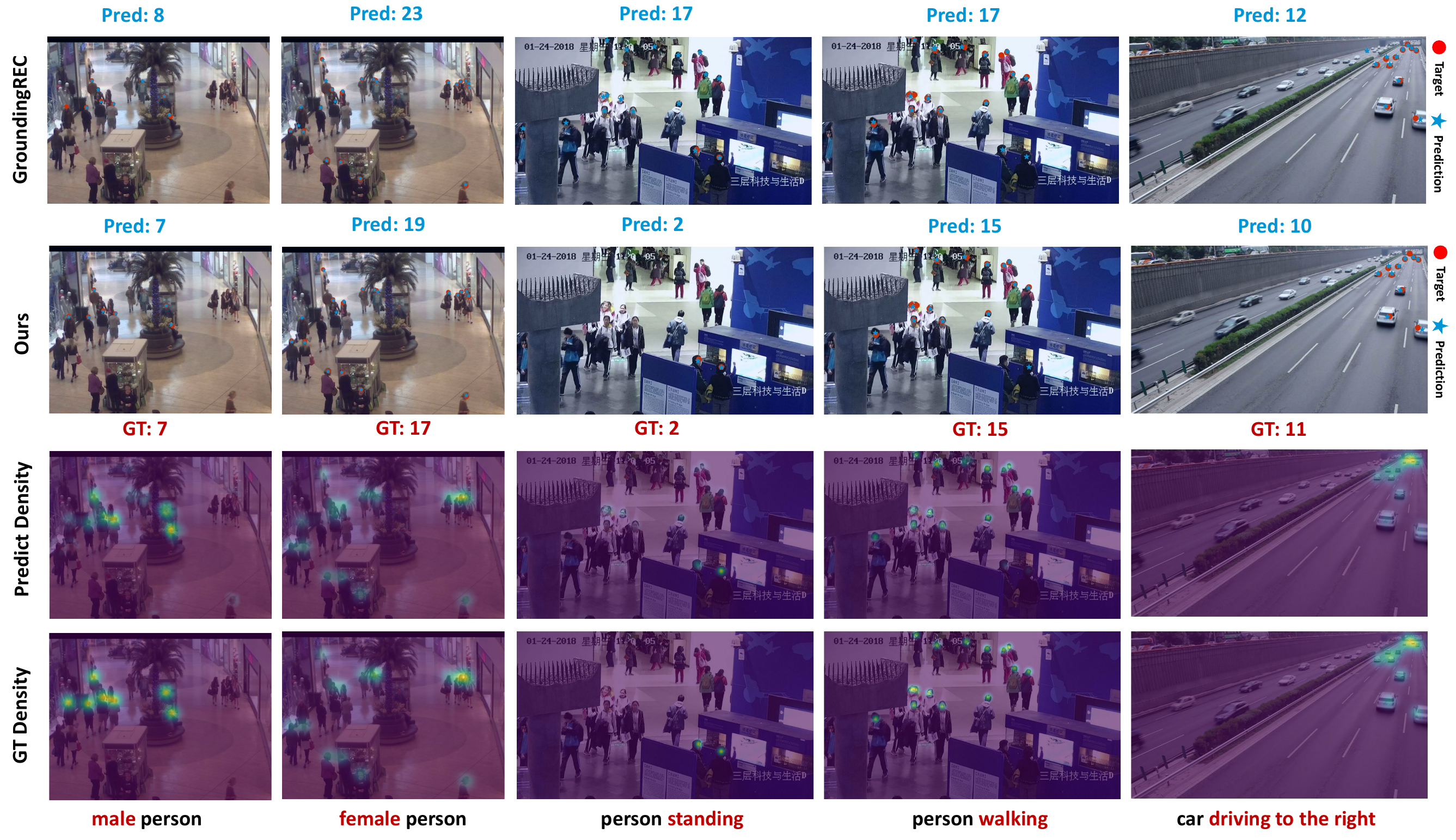}
    \vspace{-8pt}
    \caption{\textbf{Additional Qualitative Results 2.}}
    \label{fig:qualitative2}
\end{figure*}

We provide additional qualitative results of our model as Figure~\ref{fig:qualitative1} and Figure~\ref{fig:qualitative2} show.

\section{Limitations}
Due to the limitation of query quota, it is difficult to count in a query image that contains more than 900 objects. Although we can use the cropping strategy to overcome the quota, it will introduce extra computational costs.

